\documentclass[3p,compress,times,twocolumn]{elsarticle}
\usepackage{lineno}
\usepackage{amsmath}
\usepackage{amssymb}
\usepackage{caption}
\usepackage{array}
\usepackage{xcolor}
\usepackage{graphicx}
\usepackage{subcaption}
\usepackage[ruled,vlined]{algorithm2e}
\usepackage{float}
\modulolinenumbers[5]
\journal{Journal of \LaTeX\ Templates}
\begin{document}
\begin{frontmatter}
\title{Deep Neural Networks with Short Circuits for Improved Gradient Learning}

\author[mymainaddress]{Ming Yan}
\author[mysecondaryaddress]{Xueli Xiao}
\author[mymainaddress]{Joey Tianyi Zhou}
\author[mysecondaryaddress]{Yi Pan\corref{mycorrespondingauthor}}
\cortext[mycorrespondingauthor]{Corresponding author}
\ead{yipan@gsu.edu}
\address[mymainaddress]{IHPC A*STAR, Singapore}
\address[mysecondaryaddress]{Georgia State University, GA, USA}

\begin{abstract}
Deep neural networks have achieved great success both in computer vision and natural language processing tasks. However, mostly state-of-art methods highly rely on external training or computing to improve the performance. To alleviate the external reliance, we proposed a gradient enhancement approach, conducted by the short circuit neural connections, to improve the gradient learning of deep neural networks. The proposed short circuit is a unidirectional connection that single back propagates the sensitive from the deep layer to the shallows. Moreover, the short circuit formulates to be a gradient truncation of its crossing layers which can plug into the backbone deep neural networks without introducing external training parameters. Extensive experiments demonstrate deep neural networks with our short circuit gain a large margin over the baselines on both computer vision and natural language processing tasks. 
\end{abstract}

\end{frontmatter}

\section{Introduction}
Nowadays, more and more research works focus on promoting deep neural networks' performance in various aspects. From the tendency of research community, it can be categorised to: 1)transfer learning, which learns a generous representation from the large source data and transfers the learned feature to a low resource target domain (i.e. BiT~\cite{kolesnikov_big_2020}, BERT~\cite{devlin_bert_2019}); 2)neural networks architecture searching(NAS), which searches the most efficient architecture for various downstream tasks (i.e. EfficientNet~\cite{tan_efficientnet_2019}); 3)noise learning, which employs the noise boosting the neural network generality (i.e. FreeLB~\cite{zhu2019freelb}, Noisy Student Learning~\cite{chu_noisy_2020}). However, transfer learning-based methods always need large source data for pre-training. Moreover, NAS-based methods take a huge computation cost to search a high optimal network architecture on the specialized task. Furthermore, noise learning approaches need external training costs for noise learning. All the above mentioned successful works high relay on external data or computation resources. 

Besides, the shortcut connections~\cite{he2016deep} become a common component in most neural networks, which employed the residual shortcut to alleviating the gradient problem in their deep model training~\cite{tan_efficientnet_2019, devlin_bert_2019}. To further explore the shortcut, DenseNet~\cite{huang2016densely} adds the shortcut connection to every layer of its dense blob. Nevertheless, the shortcuts raise a limited promotion with a high computation and memory cost,  which strongly limits it to construct deeper neural networks and apply to wider applications. Therefore, most state-of-the-art methods still trend to build their models based on the residual shortcut, i.e., EfficientNet, Noisy Student Learning, BERT, etc. However, Veit shows the residual neural networks do not really solve the gradient problem in deep models that it only shortens the depth of deep model by the residual shortcut connections~\cite{veit2016residual}. Furthermore, the ResNet only conducts shortcuts for inner-blob without crossing-blob, which makes the gradient learning problem still exists in deep models.

As we know, the performance of deep neural networks is closely related to the 
gradient training efficiency, which conducts by continuous chain-rule multiplicative operations~\cite{glorot2010understanding, pascanu2013difficulty}. The efficient gradient learning is a consistent purchase in neural networks community. So various insightful technologies first spring out in feedforward neural networks(FNN) to boost the gradient learning, such as, weight initialization~\cite{glorot2010understanding, he2015delving}, rectifier activation~\cite{glorot2011deep, maas2013rectifier, clevert2015fast}, batch normalization~\cite{ioffe2015batch}, shortcut connections~\cite{he2016deep, huang2016densely, yu2017convolutional}, improved gradient learning~\cite{amari2000adaptive, basodi2020gradient} and gated neural networks~\cite{srivastava2015training}, etc. Moreover, most of those technologies have been introduced to recurrent neural networks(RNN) to associate the learning processing. Besides, RNN developed the specifically gradient truncation approach to improve its gradient  learning~\cite{mozer1995focused}, which is popular used in long-short term memory(LSTM) networks~\cite{hochreiter1997long} and gated neural networks~\cite{cho-etal-2014-learning}. In contrast, as far as we know, the gradient truncation of RNN is few applied to FNN.

This paper proposed a gradient enhance method, incorporated short circuit into deep neural networks, to enhance the gradient learning of backbone neural networks. The deep neural networks with SC are called short circuit neural networks(SCNet) in our paper, i.e., the backbone ResNet with SC is termed SCResNet, as well as BERT with SC is termed SCBERT. The key of SCNet is the unidirectional short circuit, which passes through the sensitive of the rear layer to the frontier layers to enhance gradient learning processing. Here, our short circuit is different from existing shortcut connections. Firstly, the short circuit is a plug-in paradigm for deep neural networks, which conducts improving gradient learning by adding a truncated gradient instead of introducing big external data or computation cost (i.e., pretraining cost in transfer learning, computation cost in noise learning and network architecture searching). Secondly, the short circuit is a unidirectional connection which propagates sensitive crosses over layers or blobs. Thirdly, our short circuit introduces the RNN gradient truncation into FNN gradient learning processing. The main contributions of our work as follows:
\begin{enumerate}
\item We proposed a short circuit connection, a unidirectional neural connection, to enhance gradient learning in deep neural networks. 
\item Short circuit introduces the gradient truncation of recurrent neural networks into feedforward neural networks without introducing external training parameters. 
\item Short circuit is a plug-in shortcut to deep neural networks. Experiments demonstrate the superiority of short circuit neural networks on both computer vision and natural language processing tasks. 
\end{enumerate}

Before elaborating our short circuit neural network, we summarize some used notations in Table~\ref{TAB:notation}: 
\begin{table}[!htbp]
\centering
\caption{Some used notations}
\label{TAB:notation}
\begin{tabular}{cl}
\hline
\multicolumn{1}{l}{\textbf{Notation}} & \multicolumn{1}{c}{\textbf{Definition}}                     \\ \hline
$x$                 & a scalar input, $X$ is vector of $x$\\
$y$                 & a scalar output, $Y$ is vector of $y$ \\
$w$                 & neural network weight , $W$ is vector of $w$ \\ 
$a^l$               & the activation of layer $l$ \\
$z^l$               & the weighted sum of layer $l$\\
$l$                 & the index of neural network layer  \\
$\delta^l$          & the sensitive of layer $l$ \\
$f$                 & the transfer function \\
$g$                 & the gate function \\ 
$L$                 & the last layer of neural network\\
$D$                 & the label or groundTruth of data \\
$J$                 & the cost function of neural network\\     \hline
\end{tabular}
\end{table}

\section{Related Works} 
\label{SEC:related_works}
\subsection{Shortcut Connections} 
Shortcut connection~\cite{he2016deep} is popular to deep model construction, which connects different layers by a shortcut connection. This manner associates the neural network to learn multiple-level features and improves model performance~\cite{zeiler2014visualizing, szegedy2015going}. It as well as releases the gradient vanishing in a deep model, extends the network to a deeper level~\cite{srivastava2015training, he2015convolutional}. Specifically, there has a variety of shortcut connections:

\paragraph{ResNet} Shortcut in ResNet~\cite{he2016deep} is conducted by an identity mapping of inputs ($a^l$, where $a^0=X$) following by Equation~\ref{eqt:resNet}. In the feedforward computation, ResNet solves the nonlinear activation function's saturation problem by enforcing $f(W^la^l)$ mapping to a none-easy-saturation zone that near to zero. In the backpropagation computation, the residual connection back propagates gradient in a shortcut that reduces the potential gradient issue alone long-distance propagation. However, the residual shortcut only exists inside of residual blobs.
\begin{equation}
\label{eqt:resNet}
a^{l+1} = f^l(W^la^l) +a^l
\end{equation}

\paragraph{HighwayNet} Highway neural network~\cite{srivastava2015training} conducts shortcut by introducing the gate function($g$) which gates its inputs(Equation~\ref{eqt:highway}). In specific, the gate function $g$ is the shortcut that controls the pass-through of nonlinear activation $f(W_fX)$ and vanilla input $X$. However, the gate of HighwayNet is a data-driven function ($g(W_gX)$) with a learnable parameter($W_g$), which increases the training cost.
\begin{equation}
\label{eqt:highway}
Y^{l+1} = f(W_fX)g(W_gX)+[1-g(W_gX)]X
\end{equation}

\paragraph{DenseNet} The shortcuts in DenseNet~\cite{huang2016densely} connect layers to each other in one dense blob. Equation~\ref{eqt:denseNet} shows the ${l+1}$ layer in DenseNet links to all frontier layers (from $i$ to $l$ layers). All the lower-level features are synthesized to the higher-level layer, which greatly improves the performance of DenseNet. Nevertheless, the quantity shortcut meanwhile leads to a high computation cost, which hinders the dense shortcut apply to more deep models.
\begin{equation}
\label{eqt:denseNet}
\begin{split}
Y^{l+1}  & =f^l(f^i, f^{i+1}, ... f^{l-1}, W^i, W^{i+1}, ... W^{l-1}, X) \\
             & = \sum_{l=1}^{l+1}\prod_{j=i}^{l}f^{j}(W^{j}Y^{j})),
\end{split}             
\end{equation}

From the above shortcut statements, deep models mostly conduct shortcuts by employing a neural connection that conducts the training process with feedforward and backpropagate computation. This bidirectional neural connection performs inside of their blobs. Moreover, all shortcuts are fixed in the model construction, even the performance of the gate function directly relates to the on-the-fly input data. So, our unidirectional short circuit is a plug-in connection to the deep neural networks without introducing external trainable parameters. 

\subsection{Gradient Truncation of Recurrent Neural Network}
The shortcut has been successfully applied to feedforward neural networks. However, the recurrent neural network differs from FNN, which is a dynamic system~\cite{funahashi1993approximation} sharing the same weight $W$ in time states($S$)(Equation~\ref{eqt:RNN}). This specialty limits the application of shortcuts in RNN. So, the RNN proposes gradient truncation to reduce the gradient problem in their long sequential patterns learning tasks. 
\begin{equation}
\label{eqt:RNN}
\begin{cases}
net(t) &= WS(t-1) + X(t-1)      \\
S(t)   &= S(t-1)  + f(net(t))   \\
Y(t)   &=  S(t),
\end{cases}
\end{equation}

Equation~\ref{eqt:RNN} shows RNN shares the same weights at different time states. Same to FNN, the gradient computation of RNN also follows the backpropagate chain-rule. To illustrate the gradient problem of RNN, we denote $\partial P(t) = \frac{\partial S(t)}{\partial W}$ (Equation~\ref{eqt:rnnGradient}). So go with time decaying, the RNN gradient easy tends to explore or vanish on the brittle condition($W{f}'(net(t-i))!=0$). Here we list the typical gradient truncation approaches are to solve this gradient problem:
 \begin{equation}
 \label{eqt:rnnGradient}
 \frac{ \partial P(t)} { \partial P(t-\tau)} = \prod_{i=1}^{\tau} (1+W{f}'(net(t-i))),
 \end{equation}
 
\paragraph{BPTT} Back propagate through time(BPTT) is a gradient-based training method that converts the long term gradient computation of RNN into a gradient computation of fix-length FNN~\cite{mozer1995focused}. In other words, the RNN unfolds into a fix-length feedforward neural network. Specifically, an RNN unroll to a $k$ layers FNN and k-steps compute the loss function. The sequential input $X_t, t\in[1,2,..,T]$ is also split into small segments($(x_1, x_2, ..., x_k), (x_{k+1},x_{k+2}), ..., $ where $k<T$ ) by the fix-length($k$). In this manner, the RNN gradient is truncated to $k$ steps. This gradient truncation operation speeds up the training efficiency of RNN. However, the fix-length unfolding manner limited BPTT to learn long-term patterns. 

\paragraph{LSTM} Long-Short Term Memory(LSTM) extends the short-term memory of BPTT to long-term memory with the gate mechanism of gradient truncation~\cite{hochreiter1997long}. Different from BPTT, LSTM cell contains input gate, memory gate, and output gate. All the gradient truncation conducts on the three types gate that truncates the gradient propagate from other LSTM cells'(Equation~\ref{eqt:lstmTruncation}). This gradient truncation manner helps the LSTM gradient flow constantly through long time steps without scaling. 
\begin{equation}
\label{eqt:lstmTruncation}
\frac{ \partial net(t)} { \partial S(t-1)}  \overset{tr}{\approx} 0,
\end{equation}

\paragraph{GRU} Gated Recurrent Unit(GRU) is a simplification of LSTM without output gate~\cite{cho-etal-2014-learning} which has a lot of variants~\cite{dey2017gate}. This simplified architecture gives GRU fewer parameters and helps it fit well on certain small datasets~\cite{Gruber2020}. But to the large-scale machine translation task, LSTM cells still consistently outperformed the GRU cells in deep models~\cite{britz-etal-2017-massive}.

Gradient truncation greatly reduces RNN gradient problem and helps RNN learning long-term patterns. Moreover, shortcut connection improves FNN gradient learning and greatly extends the depth of FNN. Shortcut connection and gradient truncation promote the gradient learning from different aspects in deep neural networks. In this paper, we proposed the short circuit to enhance the gradient learning of deep neural networks, which introduces the RNN gradient truncation into the FNN shortcut.

\begin{figure*}[!ht]
\centering
\includegraphics[width=0.95\linewidth]{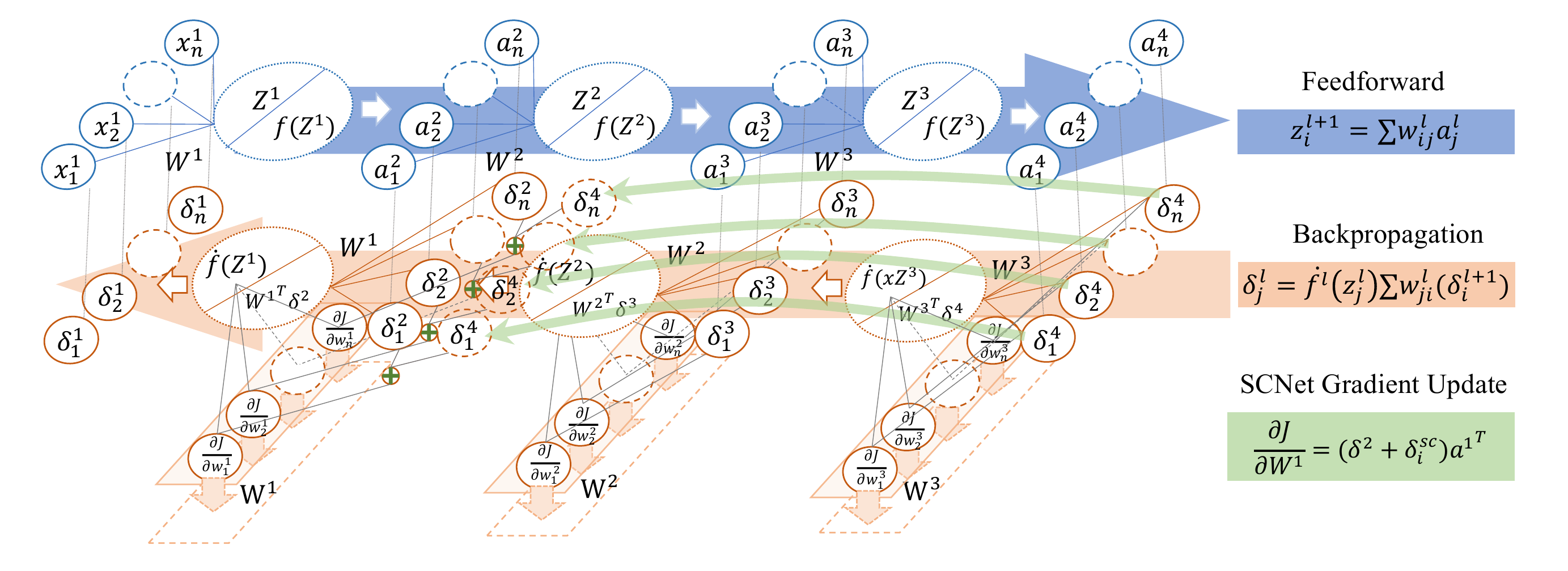} 
\caption{Overview of Short Circuit Neural Network. The blue flow denotes the feedforward computation, the orange flow denotes the backpropagation computation, and the green flow denotes the short circuit connection. This architecture illustrates the short circuit directly propagates the gradient from the third layer to the first layer.}
\label{FIG:architecture}
\end{figure*}

\section{Short Circuit Neural Network} 
Short circuit neural network(SCNet) is a neural network with the connections of short circuit.  Figure~\ref{FIG:architecture} shows an overview of SCNet. Comparing the computation to FNN, the difference of SCNet is the short circuit in green flows. In this section, we will first introduce the short circuit and its computation process. Then provides an algorithm that describes how to plug the short circuit to deep neural networks. The last is the typical applications of SCNet on specific tasks. 

\subsection{Short Circuit} 
The motivation of short circuit is to enhance the gradient in training process of deep neural network. To reducing the gradient decay in deep model, short circuit truncates the original long distance propagation route directly to the low levels. Thereby, the short circuit is a unidirectional neural connection instead of previous bi-directional shortcuts (feedforward and backpropagate computation). Here, the detailed computation of SCNet is elaborated as follows. 

Equation~\ref{Eq:feedforward1} and Equation~\ref{Eq:feedforward2} show the feedforward computation of SCNet that is the same as vanilla feedforward neural networks. Specifically, the short circuit of SCNet is a one-way neural connection in backpropagation, so there has no feedforward computation in Equation~\ref{Eq:feedforward2}:

\begin{align}
\label{Eq:feedforward1}
Z^{l+1}     &=  W^la^l\\
\label{Eq:feedforward2}
Y^{l+1}     &=  f(Z^{l+1}),
\end{align}

As well as in the backpropagate computation, the cost function ($J$) and last layer's sensitive ($\delta^L$) of SCNet is defined as:
\begin{align}
J           &= \frac{1}{2}(D-Y)^2 \\
\delta^L    &= \frac{\partial J}{\partial Z^L},
\end{align} 

Following the backpropagate algorithm, the sensitive is defined in Equation~\ref{Eq:sensitive},
\begin{equation}
\label{Eq:sensitive}
\delta^{l} = f'(Z^l)\cdot \delta^{l+1}{W^l}^T,
\end{equation}

In vanilla FNN, the gradient of $l$ layer is computed by the backpropagate algorithm(Equation~\ref{Eq:FNN_gradient}) which only receive the gradient from connected $l+1$ layer. However, in SCNet(Equation~\ref{Eq:SCNet_gradient}), the neurons in $l$ layer receive gradient both from $l+1$ layer ($\delta^{l+1} w^l{a^l}^T$) and short circuit ($\delta^{sc} w^l{a^l}^T$). Note, the neural connection actually propagates the sensitive instead of the gradient following the chain-rule of backpropagate algorithm(Equation~\ref{Eq:sensitive}), then computes the finial gradient(Equation~\ref{Eq:FNN_gradient} or Equation~\ref{Eq:SCNet_gradient}).   
\begin{align}
\label{Eq:FNN_gradient}
\frac{\partial J}{\partial W^l}  &= \delta^{l+1} {a^l}^T \\
\label{Eq:SCNet_gradient}
\frac{\partial J}{\partial W^l}  &= (\delta^{l+1} + \delta^{sc}) {a^l}^T,
\end{align}

In the last step, SCNet updates the computed gradient $\frac{\partial J}{\partial W}$ to weights$W$, as follow,
\begin{equation}
\label{Eq:update}
W := W + \frac{\partial J}{\partial W},
\end{equation}

Except for the gradient computation of short circuit layer in Equation 1, all the feedforward and backpropagate computations are the same as the traditional feedforward neural network. Generally, neural network training is a gradient optimization process of the cost function. In this training process, all the layer receives the sensitive of cost function by the chain-rule(Equation~\ref{Eq:sensitive}), then compute the gradient and update it to the weights. Typically, our short circuit connection propagates the truncated sensitive(Equation~\ref{Eq:SCNet_gradient}). Moreover, the short circuit conducts the gradient enforcement manner and relieves the long-distance gradient decay in the deep model. In this section, we introduced the training of SCNet. Next, we will introduce more details about gradient truncation in SCNet.

\subsection{Gradient Truncation of SCNet} 
In this section, we first formulate the gradient problem in feedforward neural networks~\cite{hochreiter1997long}, then provides our gradient truncation solution to this gradient problem. Our short circuit gradient truncation motive by the gradient truncation of RNN to alleviate gradient problem in FNN. Different from RNN, the FNN weights are different in each layers, which can be formulated as a function of functions from input $X$ to output $Y^L$ (Equation~\ref{eqt:mapping}):
 \begin{equation}
 \label{eqt:mapping}
 Y^{L} = f^1(W^1,f^2(W^2,...f^L(W^L,X))),
 \end{equation}
 
In the gradient computation, each layer's gradient is computed by the derivation of cost function $J$ corresponding to weight $W^l$: 
\begin{equation}
\label{eqt:gradient}
    \frac{\partial J}{\partial W^l} = \frac{\partial \frac{1}{2}(D-Y^L)^2}{\partial W^l},
\end{equation}

Following the chain-rule of backpropagate algorithm, the Equation~\ref{eqt:gradient} can extends to,
\begin{equation}
\label{eqt:gradient_extention}
   \frac{\partial J}{\partial W^l}  = (D-Y^L) \prod_{l+1}^L Y^l \cdot {f'}^l(W_lY^l),
\end{equation}

From the Equation~\ref{eqt:gradient_extention}, we observe the gradient scale is directly influenced by the term $ a^l \cdot {f'}^l(W_l a^l)$ with depth growing, where,
\begin{equation}
\label{Equ:gradient_problem}
 \frac{\partial J}{\partial W^l} = 
\begin{cases}
\infty &if \ a^l \cdot {f'}^l(w_la^l) > 1 \\
0      &if \ a^l \cdot {f'}^l(w_la^l) < 1 ,
\end{cases}
\end{equation}

The gradient vanish and exploration problem is raised by the uncertain term ($ a^l \cdot {f'}^l(W_l a^l)$), which computed by the chain-rule extension (Equation~\ref{eqt:expansion}). 
\begin{equation}
\label{eqt:expansion}
\frac{\partial J}{\partial w^l} = \frac{\partial J}{z^{L}} \frac{z^{L}}{\partial w^l} = \frac{\partial J}{\partial Z^{L}} \frac{\partial Z^L}{\partial Z^{L-1}} 
                    \frac{Z^{L-1}}{Z^{L-2}} \cdots \frac{Z^{l+1}}{Z^l} \frac{\partial Z^l}{\partial W^l},
\end{equation}

To relieve the gradient problem in deep model, SCNet truncate the chain-rule computation to be a constant value(Equation~\ref{eqt:truncation}) from $L-(m+1)$ layer to layer $l=n$ layer. In this manner, the short circuit connection directly propagate the sensitive ($\delta^{sc}$ from rear layer back to the front layer(Equation~\ref{Eq:sensitive}). 
\begin{equation}
\label{eqt:truncation}
    \prod_{l=n}^{L-m-1} \frac{\partial Z^{l+1}} {\partial Z^l} \overset{tr}{\approx} 1.
\end{equation}

SCNet truncates the internal computation of chain-rule to constant one that constrained the gradient scaling in deep neural networks. Moreover, the short circuit connection directly propagates the sensitive cross multiple layers and conducts a deep model gradient enhancement. Typically, SCNet has two differences from the gradient truncation of RNN. Firstly, the RNN gradient truncation truncates the gradient of outside-cell to be zero, but our SCNet truncates chain-rule internal computation to be one. The other difference reflects on the connections where the connection of RNN is a bidirectional link with feedforward and backpropagate computation. In contrast, our short circuit is a unidirectional connection without feedforward computation.

\subsection{Algorithm of SCNet}
There are two precondition for the short circuits: 1). the front layer ($l_i, i\in N^+$) and rear layer($l_{sc}$) in short circuit connection must have the same neurons; 2). the index of rear layer should larger than the index of front layer ( $i < {sc}$).

\begin{algorithm}
\SetAlgoLined
\KwIn{ {$(X,Y):$} training data;\\ {$l_{sc}:$} rear layer of SC;\\ {$k:$} skipping layers of SC.\\}
\KwResult{SCNet.}
Initialize backbone weights {$W^l$}\;
\For{i = 1:Epochs} {
        \For{Sample {$(x,y)$} from {$(X,Y)$}}{
            Feedforward Computation\;
            \For{l=1:L}{
               {$Z^{l+1} = W^la^l$}\;
               {$a^{l+1} = f^{l+1}(Z^{l+1})$}\;
            }
            Backpropagate Computation\;
            \For{l=L:1}{
                \eIf{l==L}{
                     { $\delta^L = f'^L(Z^L)\cdot(Y-a^L)$ }\;
                }{
                     { $\delta^l = f'^l(Z^l)\cdot\delta^{l+1}{W^l}^T$ }\;
                }
                \eIf{$mod(l,k)==0$ {\bf and} $l<sc$}{
                     $\frac{\partial J}{\partial W^l}  = (\delta^{l+1} + \delta^{sc}) {a^l}^T$ \;
                }{
                     $\frac{\partial J}{\partial W^l}  = \delta^{l+1} {a^l}^T$ \;
                }
           }
           $W =: W + \frac{\partial J}{W}$\;
  }
}
\caption{Algorithm of SCNet}
\label{Al:SCNet}
\end{algorithm}

\begin{figure*}[!htp]
\centering
\includegraphics[width=0.9\textwidth]{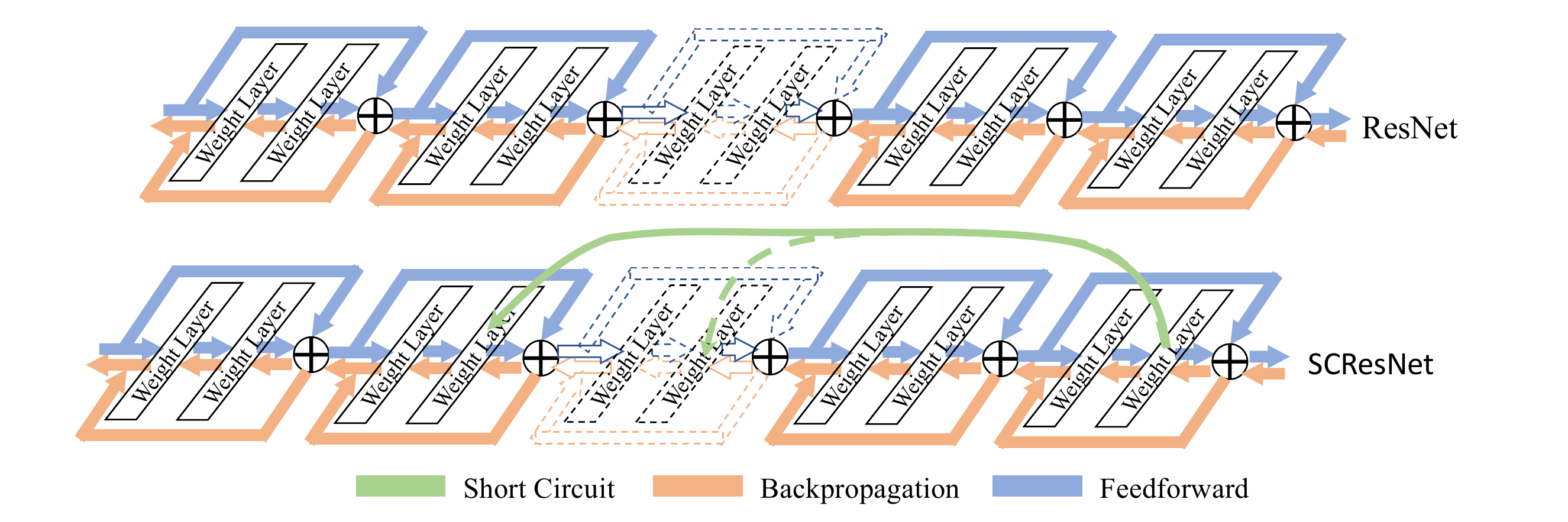} 
\caption{ResNet and SCResNet Comparison.}
\label{FIG:SCResNet}
\end{figure*}

Algorithm~\ref{Al:SCNet} shows the algorithm of SCNet that consists of the feedforward and backpropagation parts. The $l_{sc}$ denotes the rear layer. $K$ denotes the gap of skipping layers between the rear layer and front layers. From the algorithm, we observe the short circuit has no computation in the feedforward computation and only exist in the backpropagate computation. The single SC gradient is added only by predefined $k$ layers. Our algorithm illustrates the process of SCNet, which can also extend to multiple versions, such as the specific SC rear layer can extend to multiple versions, and different strategies can introduce to SC building. Following the two preconditions, short circuit connections can generalize to diverse variants on different backbone models and downstream tasks.

\subsection{Application of SCNet}
\label{Section:Application_SCNet}
Short circuit neural networks focus on gradient-based enhancement by incorporating short circuit connections into backbone neural networks. Mostly, those backbone neural networks are the leading approaches for different tasks. So, SCNet stands the shoulder of those backbone methods and enhances gradient learning of those backbones by adding short circuit connections. 

ResNet is a popular neural architecture in computer vision tasks. Thereby, we apply the short circuit to ResNet, which is termed SCResNet in this paper. Figure~\ref{FIG:SCResNet} illustrates a comparison between backbone ResNet and SCResNet. Short circuit connection does not change the network architecture of the backbone. The only difference is our SCResNet has the external back direction link, which propagates the sensitive from the rear layer to the front layers. Moreover, this external shortcut enhances gradient learning with a low cost (Equation~\ref{Eq:sensitive} than other popular approaches( i.e. adversarial training, NAS~\cite{JMLR:v20:18-598, xiao2019fast}).

\begin{figure*}[!htp]
\centering
\includegraphics[width=0.95\textwidth]{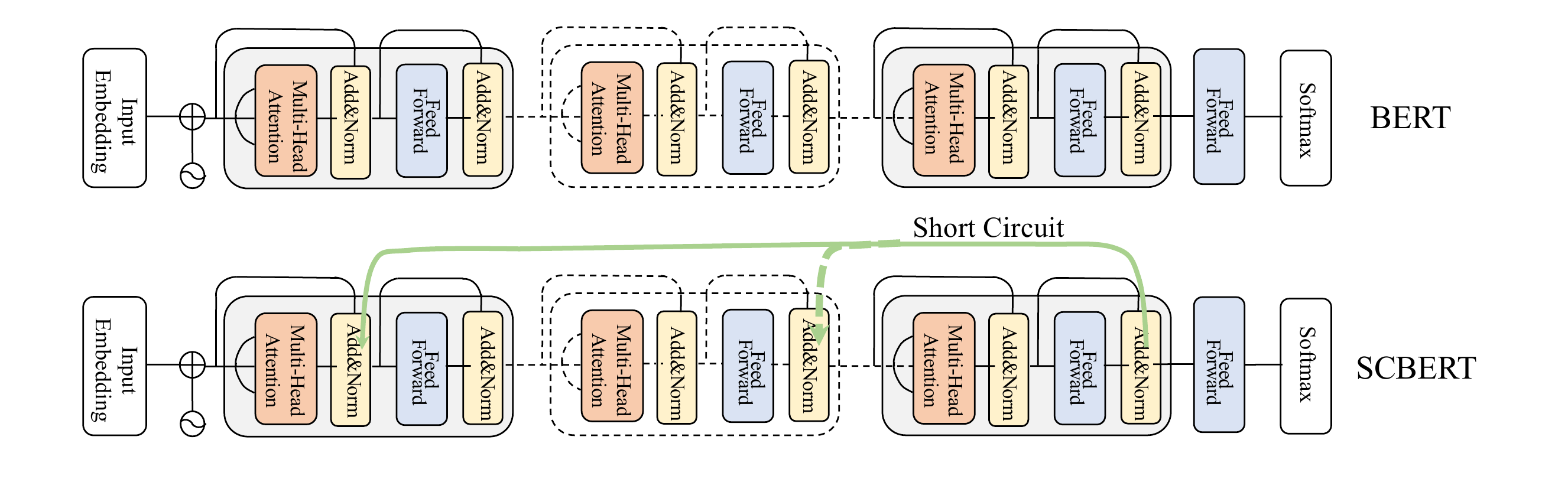} 
\caption{BERT and SCBERT Comparison.}
\label{FIG:SCBERT}
\end{figure*}

As well as SCResNet, we also apply the short circuit connection to BERT. BERT is consists of transformer encoder blob, which is fully connected network with residual shortcuts.Figure~\ref{FIG:SCBERT} illustrates the network structure of vanilla BERT and SCBERT. The short circuit connection extends the in-blob shortcut to crossing-blob shortcut that propagates the sensitive cross BERT encoder blobs in SCBERT.


\section{Experiments}
In this section, we first briefly introduce the datasets in our experiments, then apply the short circuit to different backbones: ResNet, BERT, and Roberta to be SCResNet, SCBERT, and SCRoberta. At last, evaluate the performance of those short circuit neural networks on different tasks. 

\subsection{Dataset}
For the computer vision task, we mainly evaluate our SCResNet on the CIFAR10 that is a subset of tiny images dataset~\cite{krizhevsky2009learning}. The CIFAR10 dataset consists of $60,000$ color images with the resolution of $32\times32\times 3$. The test dataset has $10,000$ images for ten classes.  We also introduced the MNIST dataset~\cite{lecun1998gradient} in the discussion section. MNIST contains $10$ classes gray handwriting digit images. Each category has $6000$ images with a fixed resolution of $28\times28$. The rest $10,000$ images are used for the test dataset.

For the natural language processing task, we evaluate our SCBERT and SCRoberta on the multiple choice reading comprehension question answering (MCQA) datasets: DREAM and SemEval-2018. Different from the computer vision dataset, the MCQA sample consists of context, question, and candidate choices. Specifically, DREAM's context type is the dialogue with 6444 training samples, and each question has four candidate choices. While the context type of SemEval-2018 is narrative text with 2119 samples, and each question has two candidate choices. 

\subsection{Performance on CIFAR10}
Following the Algorithm~\ref{Al:SCNet}, we first applied the short circuit connection to the backbone of ResNet-V1~\cite{he2016deep}. The reason we choice ResNet-V1 is that its performance get declined in the deep model. Figure~\ref{FIG:SCResNet} shows the architecture of SCResNet with the unidirectional SC connections. SC propagate the sensitive(green flows) of rear layer to the front layers, then calculate the truncated gradient of short circuit. To evaluate the performance of SCResNet, we test the performance of SCResNet on the CIFAR10 dataset. 

\begin{table}[!htbp]
\centering
\begin{tabular}{lcc}
\hline
\textbf{Names} & \textbf{ResNet} & \textbf{SCResNet} \\ \hline
Layer-20    & 91.25 & 91.81 \\
Layer-32    & 92.49 & 92.94 \\
Layer-44    & 92.83 & 93.29 \\
Layer-56    & 93.03 & 93.81 \\
Layer-110   & 93.39 & 94.13 \\
Layer-1202  & 92.07 & 94.46 \\ \hline
\end{tabular}
\caption{Comparison of ResNet and SCResNet on CIFAR10.}
\label{tab:ResNet_Comparison}
\end{table}

The results of SCResNet and ResNet is summarized in Table~\ref{tab:ResNet_Comparison}. From the results, we observe the performance of baseline ResNet improves with the model depth increasing. However, once the depth excesses $1K$ layers,  the increasing performance gets a sharp decline, which almost back to ResNet56. One possible reason is the gradient learning decline in the deep sub-residual block(about 300 layers). In contrast, our SCResNet keeps a consistent promoting with dept increasing from layer-20 to layer-1202. Short circuit associate SCResNet got better performance in all different depths. To the deepest layer-1202 model, the short circuit even boosts $2.4\%$ performance than the baseline ResNet. 


\subsection{Performance on DREAM}
For the natural language processing task, we employ the popular language models BERT and Roberta as the backbones for the multiple choice question answering task. Following the Algorithm~\ref{Al:SCNet}, short circuit plugs into backbone BERT to be the SCBERT, as well as the stronger baseline Roberta to be the SCRoberta. The short circuit skipping gap is set $k=4$, the rest experiments setting follows ~\cite{devlin_bert_2019}.

\begin{table}[!htp]
\centering
\begin{tabular}{lcll}
\hline
\multicolumn{1}{l}{\textbf{Names}} & \textbf{Models} & \multicolumn{1}{c}{\textbf{Dev}} & \multicolumn{1}{c}{\textbf{Test}} \\ \hline
SAR~\cite{chen-etal-2016-thorough}      & -     & 40.2 & 39.8 \\
GAR~\cite{dhingra-etal-2017-gated}      & -     & 40.5 & 41.3 \\
Co-Matching~\cite{wang-etal-2018-co}    & -     & 45.6 & 45.5 \\
FTLM~\cite{radford2018improving}        & -     & 55.9 & 55.5 \\
XLNet~\cite{yang2019xlNet}              & Large & -    & 72.0 \\
BERT~\cite{Sun2019Probing}              & Base  & 63.2 & 63.2 \\
BERT~\cite{Sun2019Probing}              & Large & 66.0 & 66.8 \\
Roberta~\cite{perez-etal-2019-finding}  & Large & 85.4 & 85.0 \\ \hline
SCBERT                                  & Base  & 63.3 & \textbf{63.3} \\
                                        & Large & 66.6 & \textbf{67.6} \\
SCRoberta                               & Large & 87.5 & \textbf{86.3} \\ \hline
\end{tabular}
\caption{Performance on Dream.}
\label{tab:SC_DREAM}
\end{table}

Table~\ref{tab:SC_DREAM} reports the comparison of our SCNets to other SOTA baselines. From the numbers, we observe the performance of our SCRoberta outperform a large margin than the none-SCNet baselines(i.e., XLNet, BERT, Roberta) and none-pretrained methods(i.e. FTLM, Co-Matching). Typically, the large-size models get more improvements than the base-size model. In other words, the short circuit is more fit for the deep models. One straightforward reason is that the larger ones get more short circuit connections (Algorithm\ref{Al:SCNet}) and strong capability than the base one. 

\subsection{Performance on SemEval-2018}
We further evaluate the effective of short circuit on SemEval-2018 where the answer candidates less than that of DREAM. Experiments setting as well as DREAM, except the downstream classifier need to fit the two candidate choices in SemEval-2018. 

\begin{table}[!htp]
\centering
\begin{tabular}{lccc}
\hline
\multicolumn{1}{l}{\textbf{Names}} & \textbf{Models} & \textbf{Dev} & \textbf{Test} \\ \hline
MITRE~\cite{merkhofer-etal-2018-mitre}      & -         & 85.1  & 82.3\\
ConceptNet~\cite{wang-etal-2018-yuanfudao}  & -         & 85.3  & 83.9 \\
GPT~\cite{sun-etal-2019-improving}          & Base      & 84.1  & 88.0 \\
GPT~\cite{sun-etal-2019-improving}          & Large     & -     & 88.6 \\
GPT~\cite{sun-etal-2019-improving}          & Large$^+$ & -     & 89.5 \\
BERT~\cite{10.1145/3357384.3358165}         & Base      & -     & 87.53 \\
BERT~\cite{jin2019mmm}                      & Large     & -     & 88.7 \\
Roberta~\cite{yan-etal-2020-multi}          & Large     & 93.76 & 94.0 \\ \hline
SCBERT                                      & Base      & 88.0  & \textbf{88.1} \\
                                            & Large     & 88.9  & \textbf{89.2} \\
SCRoberta                                   & Large     & 94.8  & \textbf{94.7} \\ \hline
\end{tabular}
\caption{Performance on SemEval-2018, where `$^*$' denotes a large-size GPT with strategies.}
\label{tab:SC_SemEval2018}
\end{table}

The results of SemEval-2018 are summarized in Table~\ref{tab:SC_SemEval2018}. From the results, we can first observe that the short circuit connection also promote the performance of backbone models. Due to the original high performance of baselines, the promotion of SCBERT and SCRoberta is less that on DREAM. Moreover, the large-size model also gains more in their performance than the base-size model.

\section{Discussions}
In this section, we further explore more about short circuit on training efficiency, gradient comparison, and parameter sensitive. Due to BERT and Robert's high computation cost, we mainly focus the discussion on ResNet and SCResNet. 

\subsection{Efficiency of SCNet}
\begin{figure}[!htp]
\centering
\includegraphics[width=0.75\linewidth]{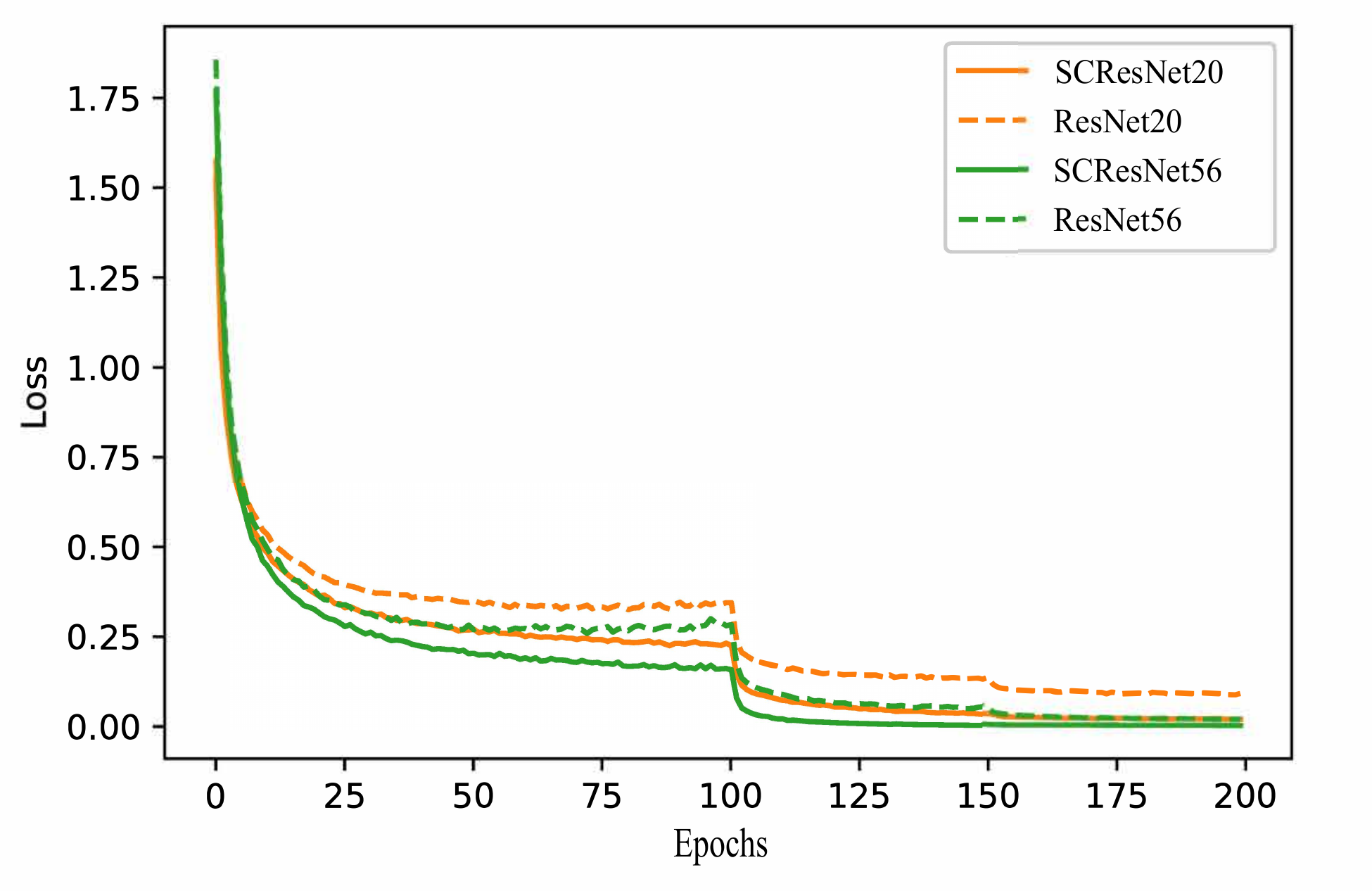} 
\caption{Training Comparison on CIFAR10.}
\label{FIG:comparison_eff_cifar10}
\end{figure}
Figure~\ref{FIG:comparison_eff_cifar10} shows the loss convergence comparison between SCResNet and ResNet on layer-20 and layer-56. From the comparison we observe that the residual networks with the short circuits get a better convergence than the baseline models. SCResNet with the short circuit get a consistent fast convergence than ResNet on the training loss. Thereby, SCResNet got a good performance on the gradient-base learning(Table~\ref{tab:ResNet_Comparison}).

\subsection{Gradient Analysis}         
Except for the neural network loss, the gradient is another signification signal for neural network learning. So, we explore the internal gradient changes in the short circuit neural networks. To better illustrate the gradient problem in the deep neural network, we employ a fully connected neural network to illustrate how our short circuit enhances gradient learning.  Then, a detailed gradient comparison is discussed between ResNet and SCResNet on different depths and training periods. 

\subsubsection{Gradient Analysis on Fully Connected Neural Network} 
\label{sec:FCN}

\begin{figure}[!htp]
\centering
\includegraphics[width=.8\linewidth]{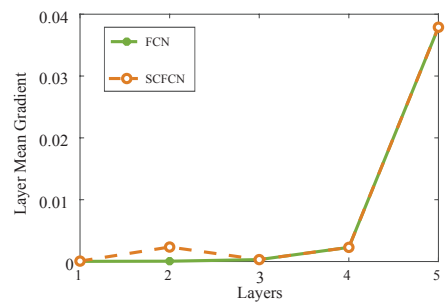}  
\caption{Mean Gradient Comparison on MNIST. The mean gradient denotes the mean of the neurons' gradient in a certain layer.}
\label{FIG:comparison_eff_MNIST}
\end{figure}

\begin{figure*}[!htp]
\centering
\begin{subfigure}{.32\linewidth}
  \includegraphics[width=\linewidth]{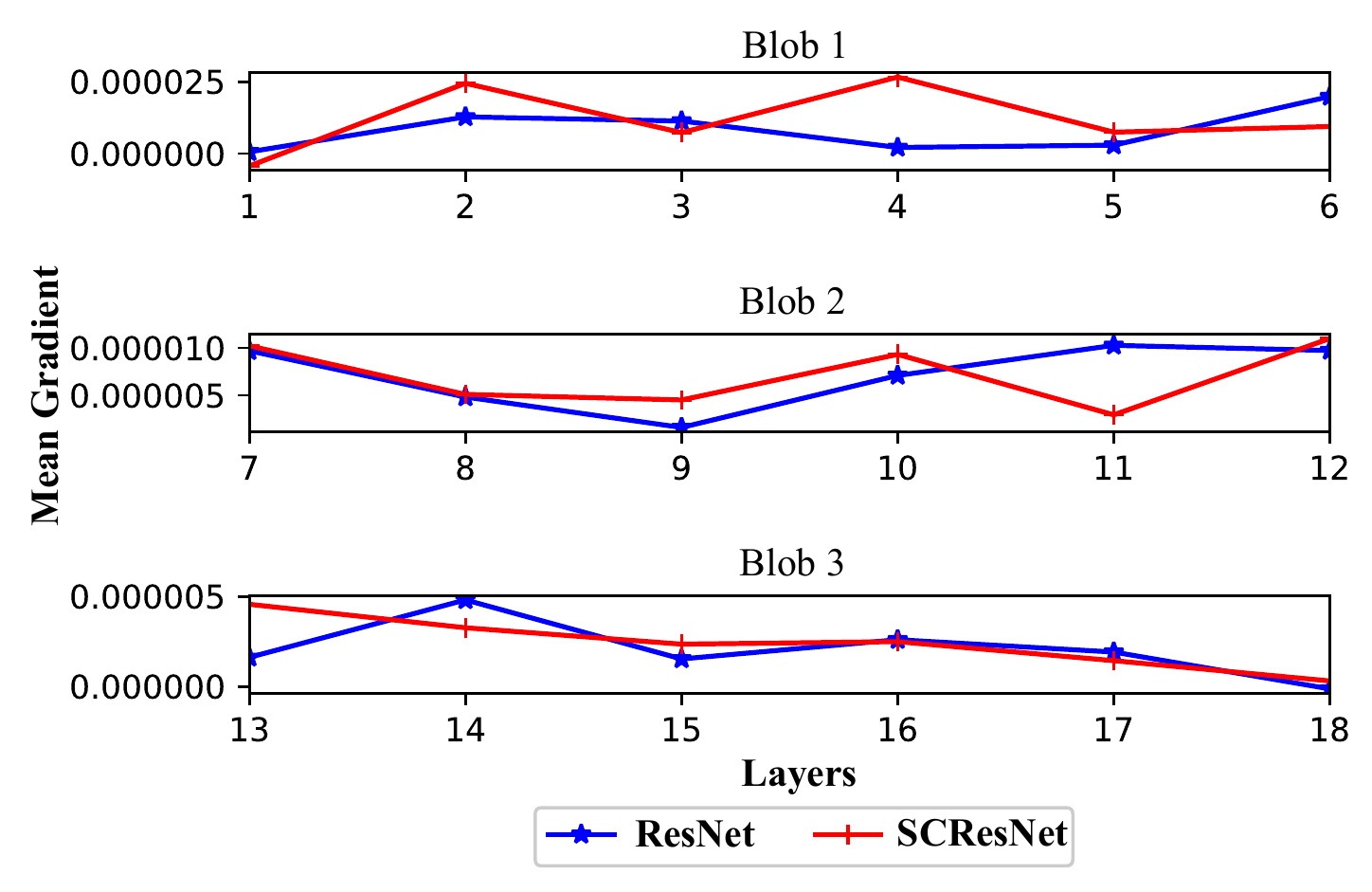}
  \caption{Epoch 1 }
  \label{fig:sfig1}
\end{subfigure}%
\begin{subfigure}{.32\linewidth}
  \includegraphics[width=\linewidth]{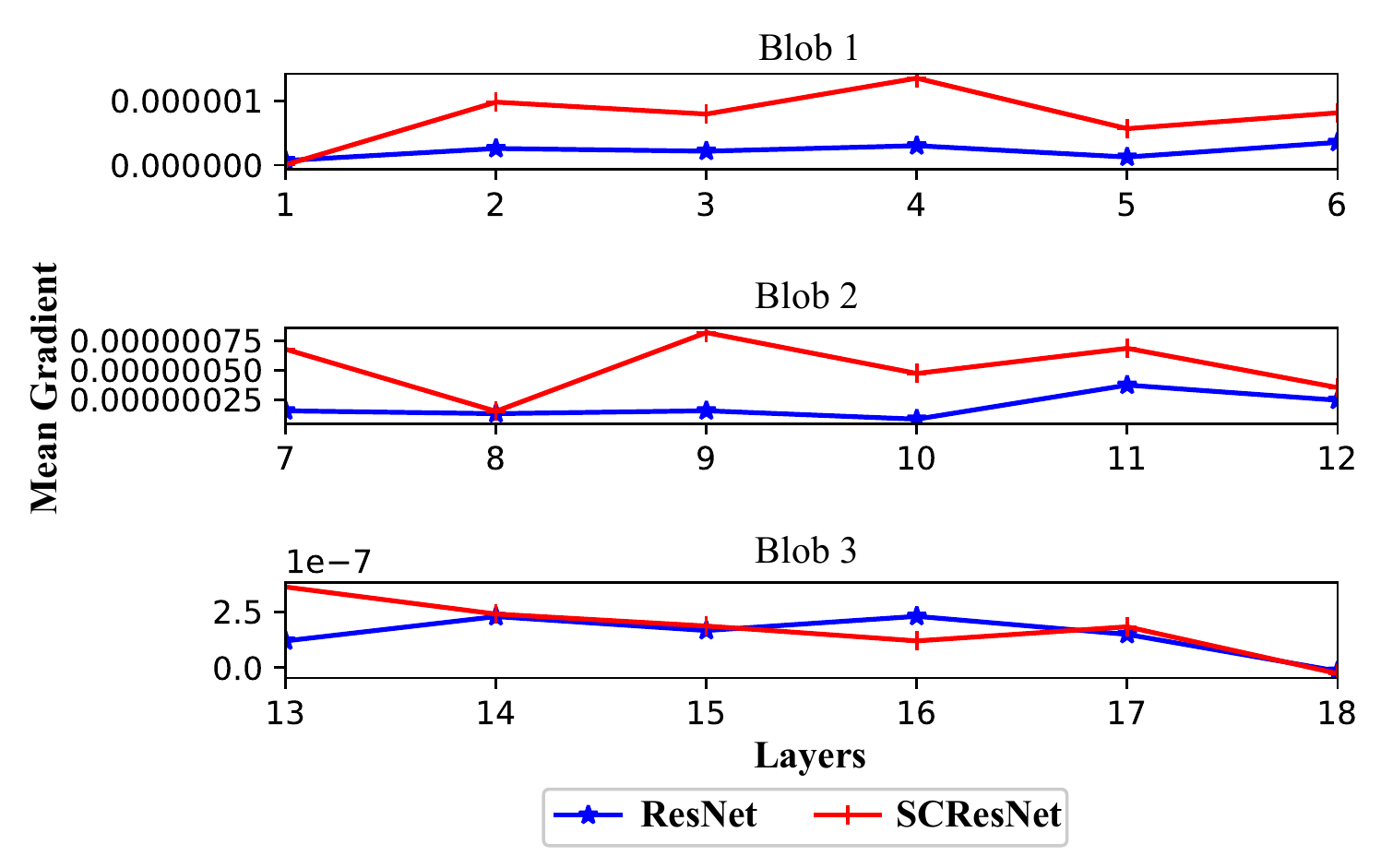}
  \caption{Epoch 1-100 }
  \label{fig:sfig2}
\end{subfigure}
\begin{subfigure}{.32\linewidth}
  \includegraphics[width=\linewidth]{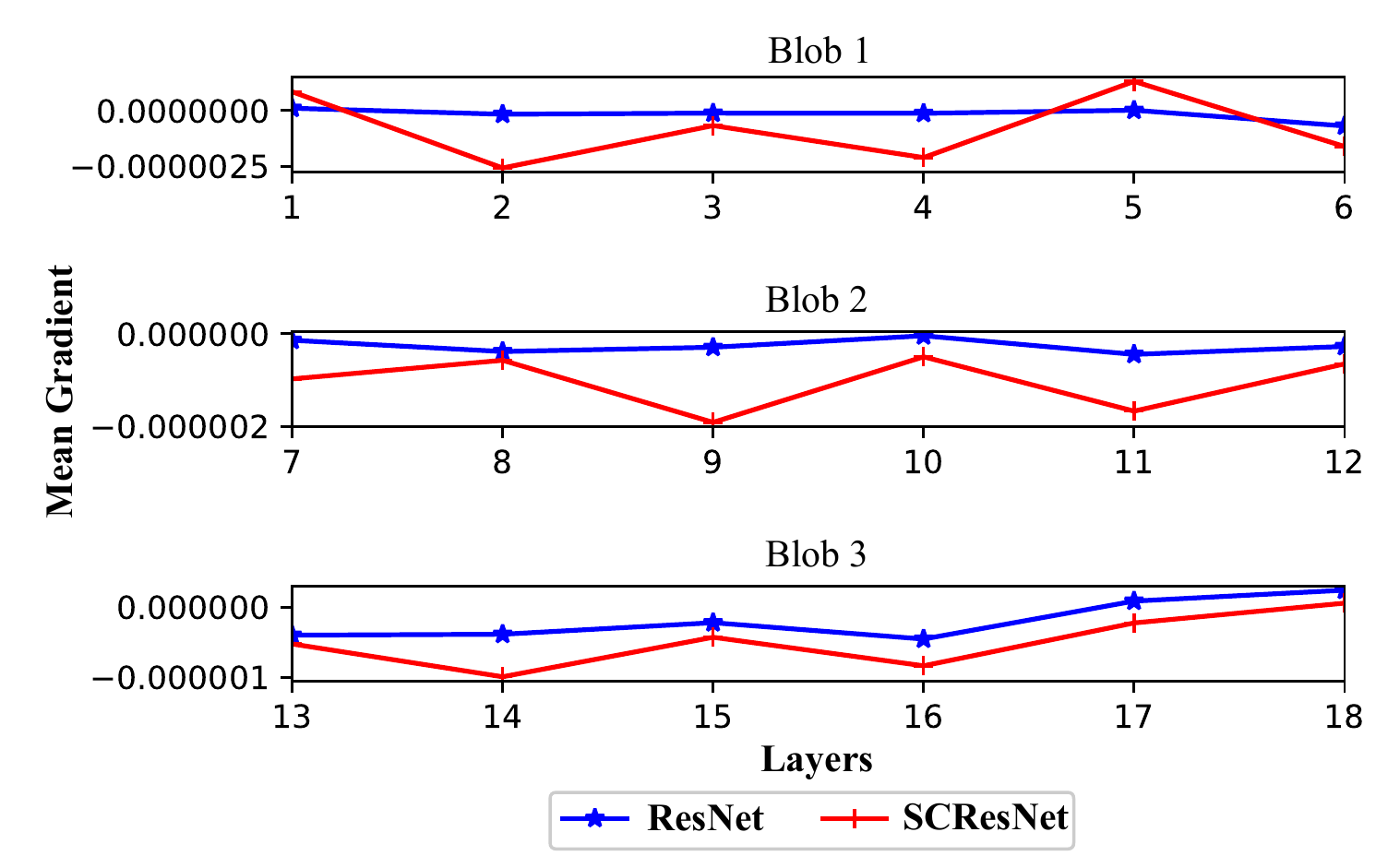}
  \caption{Epoch 1-200 }
  \label{fig:sfig3}
\end{subfigure}%

\begin{subfigure}{.32\linewidth}
  \includegraphics[width=\linewidth]{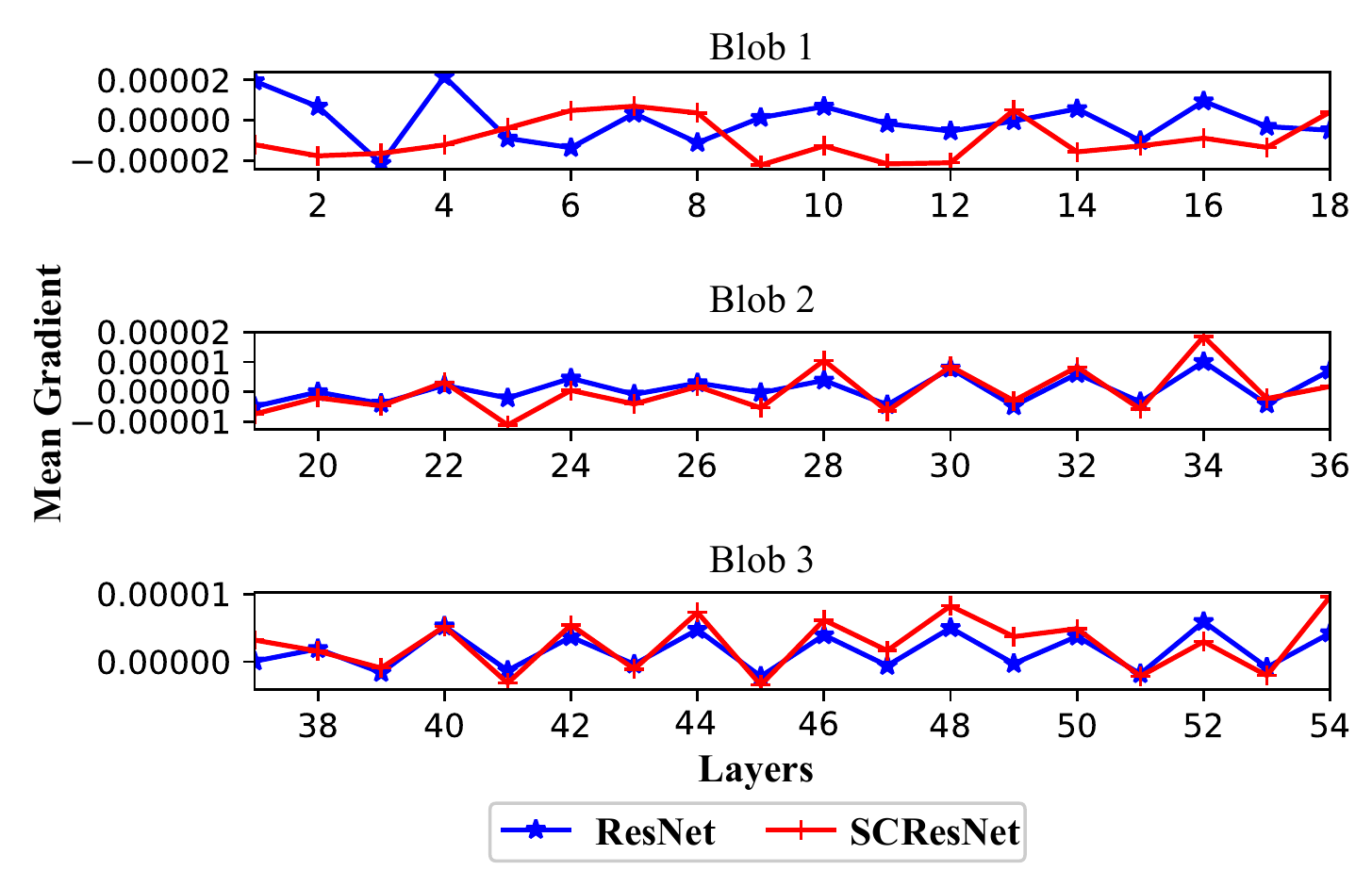}
  \caption{Epoch 1 }
  \label{fig:sfig4}
\end{subfigure}%
\begin{subfigure}{.32\linewidth}
  \includegraphics[width=\linewidth]{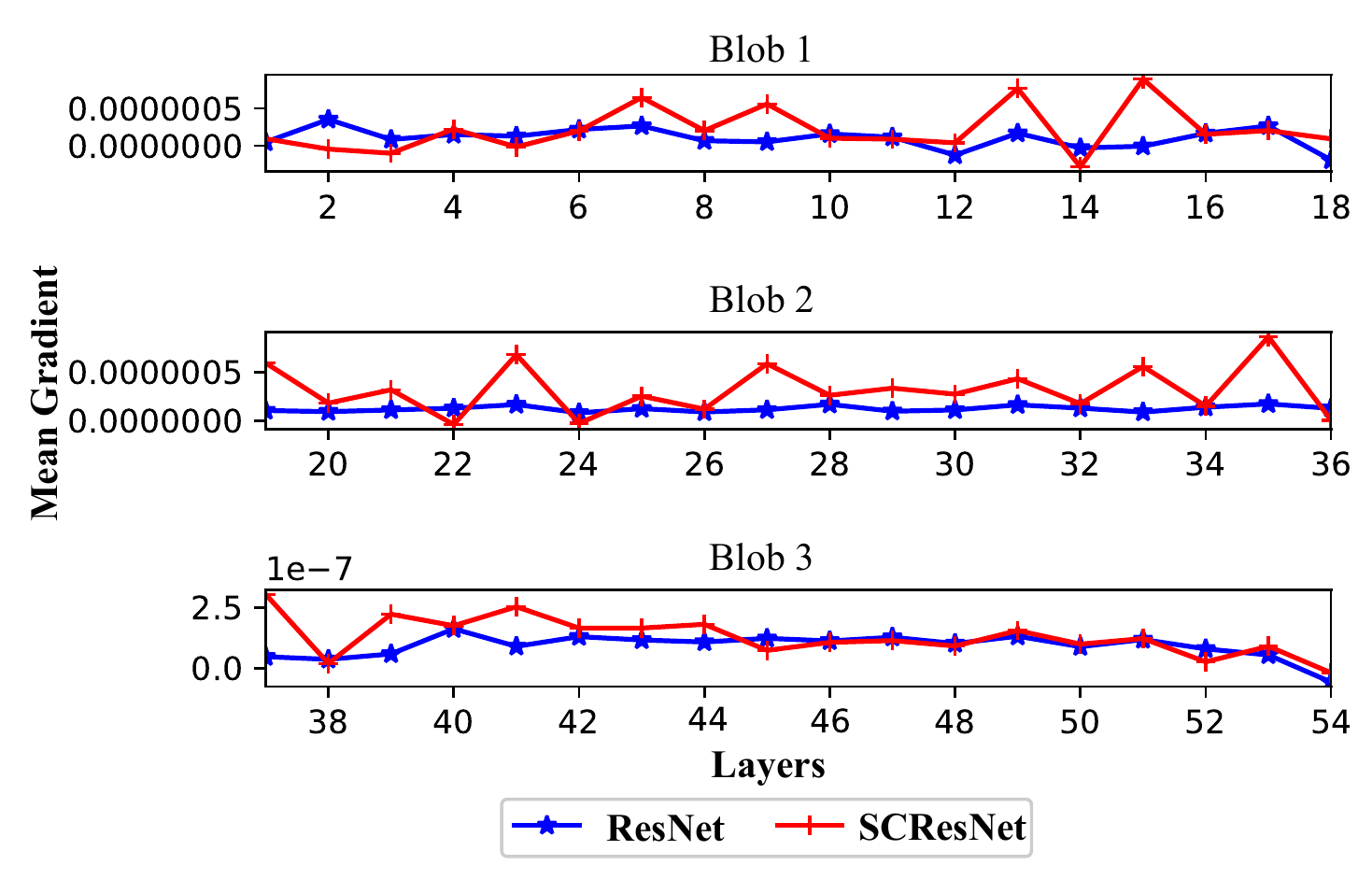}
  \caption{Epoch 1-100}
  \label{fig:sfig5}
\end{subfigure}
\begin{subfigure}{.32\linewidth}
  \includegraphics[width=\linewidth]{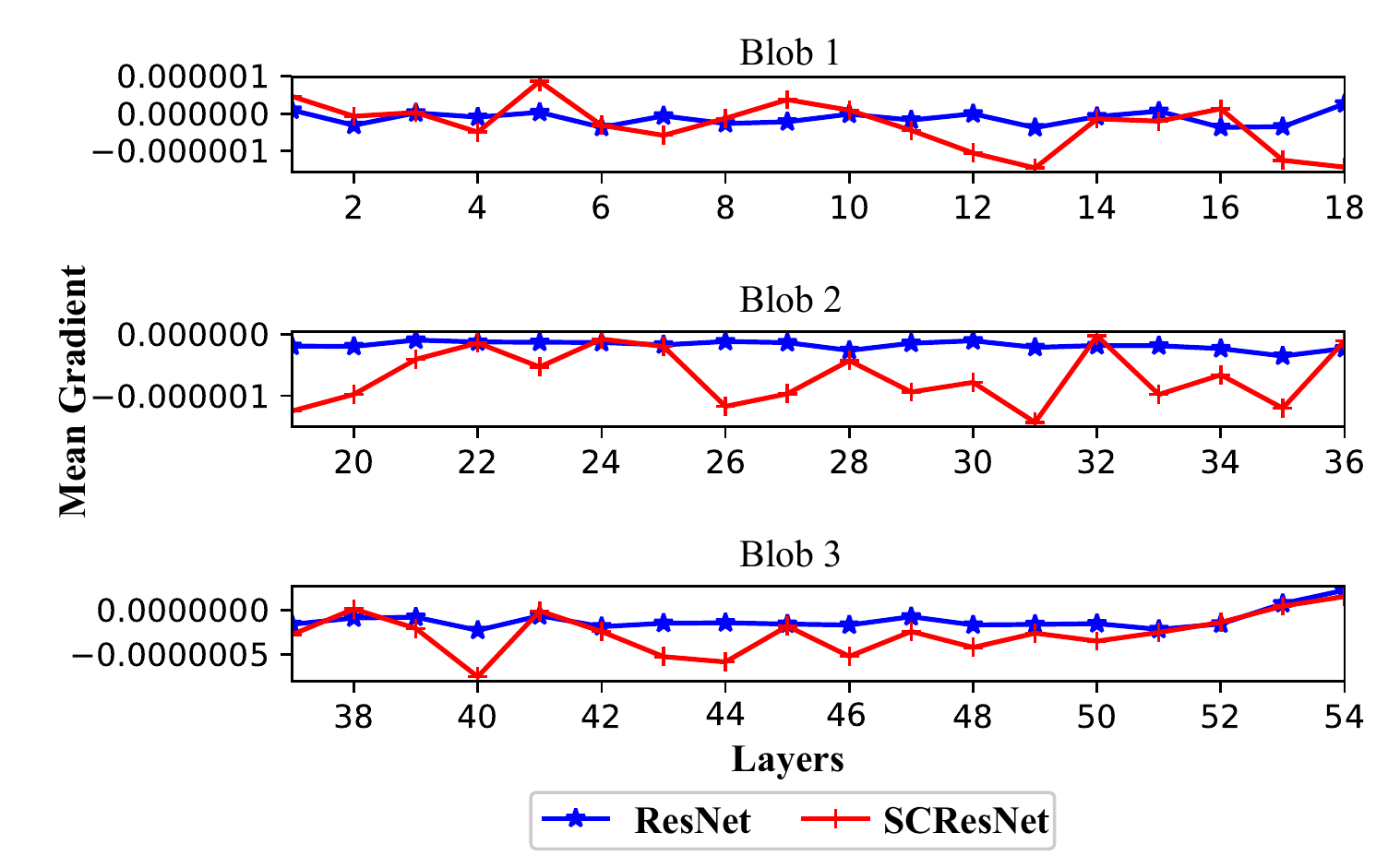}
  \caption{Epoch 1-200}
  \label{fig:sfig6}
\end{subfigure}%
\caption{The Mean Gradient Comparison of ResNet and SCResNet. The X-axis is the layer index, the Y-axis is the mean gradient of different layer, and each sub-figure contains three residual blobs.  Figure~\ref{fig:sfig1} - Figure~\ref{fig:sfig3} is the mean gradient comparison of ResNet20 and SCResNet20 in different training period, as well as Figure~\ref{fig:sfig4} - Figure~\ref{fig:sfig6} is the comparison of ResNet56 and SCResNet56. Taking Figure~\ref{fig:sfig3} as an example, `Epoch 1-100' denotes the mean value of gradient in different layers from epoch 1 to epoch 200.}
\label{fig:fig}
\end{figure*}

To illustrate the typical gradient problems in deep neural networks, we configure five layers fully connected neural network(FCN). And the activation function is set to Sigmoid function, which typically presents the gradient problems with layer increasing. For the comparison, the short circuit is added from the fourth layer to the second layer, which is the short circuit fully connected network(SCFCN).

The mean gradient comparison of FCN and SCFCN is plotted in Figure~\ref{FIG:comparison_eff_MNIST}. From the mean gradient curves, we observe the gradient decline sharply alone backpropagation flow that the mean gradient value scaled from $1e-2$ down to $1e-4$ in only five FCN layers. In contrast, the short circuit connection in SCFCN greatly enhanced the gradient on the second layer from the fourth layer in a manner of gradient truncation. 

\subsubsection{Gradient Analysis on ResNet}
We further analyze the training gradient transformation of ResNet and SCResNet on different layers. As well as the previous FCN setting, ResNet and SCResNet also employed the mean gradient in their gradient analyzation. While the difference is the residual networks have more complex network architectures and layers than the FCN. 


Figure\ref{fig:sfig1} - Firgure\ref{fig:sfig3} are the mean gradient comparison between ResNet and SCResNet on depth 20. Most mean gradient values are positive in the first training epoch on both ResNet and SCResNet (Figure\ref{fig:sfig1}). And the second and fourth layer gradient on the first blob are enhanced by short circuit connections in SCResNet20. Till to the training epoch 100 (Figure\ref{fig:sfig2}), the gradient in first and second blobs are significant enhanced by the short circuit connections. With the training processing to epoch 200 (Figure\ref{fig:sfig3}), more and more layers are dominated by the negative gradient that most mean gradient becomes to negatives. In contrast to the positive values in Figure\ref{fig:sfig2}, the enhanced layers still keeps a large mean gradient response. 

With the model depth goes to deeper, the mean gradient in ResNet56 and SCResNet56 (Figure\ref{fig:sfig4} - Figure\ref{fig:sfig6}) become more complex than the ones at 20 layers(Figure\ref{fig:sfig1} - Figure\ref{fig:sfig3}). In the first epoch (Figure\ref{fig:sfig4}), we observe only the first blob exists large difference on their mean gradient. Until the training middle stage (Figure\ref{fig:sfig5}), the significant gradient difference still only reflect on the specific short circuit connection layers. As well as Figure\ref{fig:sfig3}, Figure\ref{fig:sfig6} shows the whole training process of ResNet56 and SCResNet56 are updated by the negative gradient.

\subsection{Different Short Circuit Connections}

\begin{figure}[!htb]
\centering
\includegraphics[keepaspectratio,width=0.8\linewidth,height=6cm]{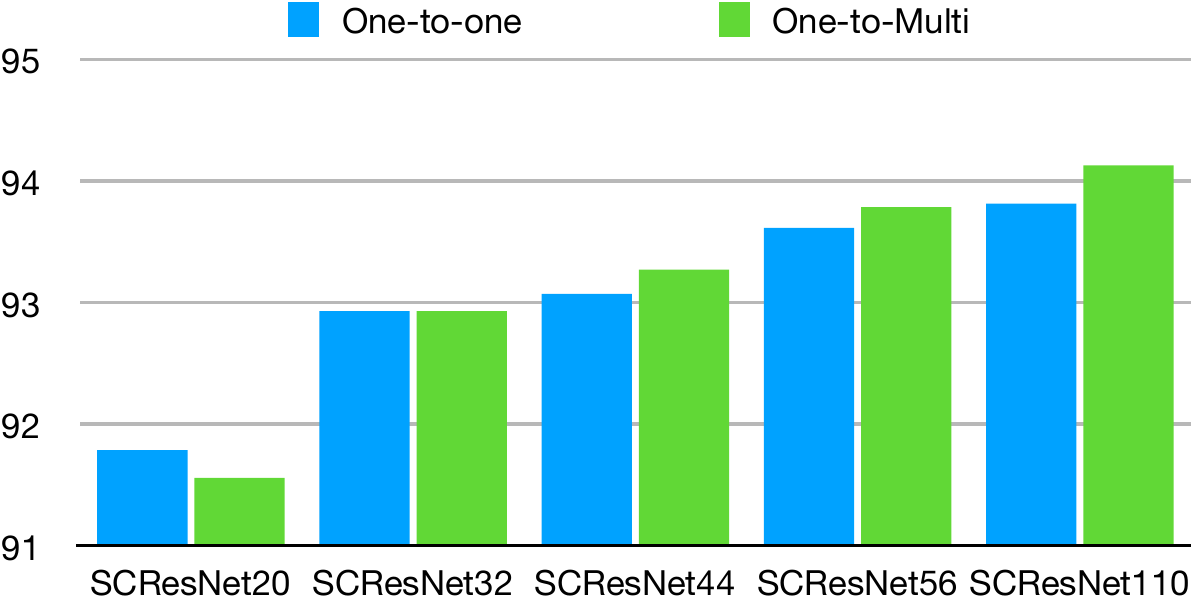} 
\caption{Different Connections of Short Circuit.}
\label{FIG:comparison_connection}
\end{figure}

We explored the relationship between short circuit connections and the performance on different backbone models. From the results of Figure~\ref{FIG:comparison_connection}, we learn multiple short circuit connections do not always get better performance. For example, the multiple SC connections of SCResNet32 almost got no improvement. Furthermore, the performance of SCResNet20 even declines after plugged multiple SC connections. However, the results got to converse on the deep models. Multiple SC connections benefit more to the performance of deep models than the single SC connection. With the depth increases, the results show the performance boosts more. The multiple SC connections might enhance the gradient in a deep model while increasing the learning complexity in the shallow models. 

\subsection{Parameter Sensitive}
This section conducts sensitive experiments on short circuit connection: short circuit weight and training batch size. Gradient decay is a common problem in deep models, and two ends of short circuit connections always have different scales in gradient learning (see Section~\ref{sec:FCN}). To investigate the scale effect for short circuit, we set the short circuit weight to be an adaptor for the different gradient scales in the short circuit shortcut. Meanwhile, the batch size is an impact factor for the batch normalization, which becomes a common component in deep models. In this section, we also explore the effects of batch size in short circuit neural networks.

\begin{figure}[!htp]
\centering
\includegraphics[keepaspectratio,width=0.8\linewidth,height=6cm]{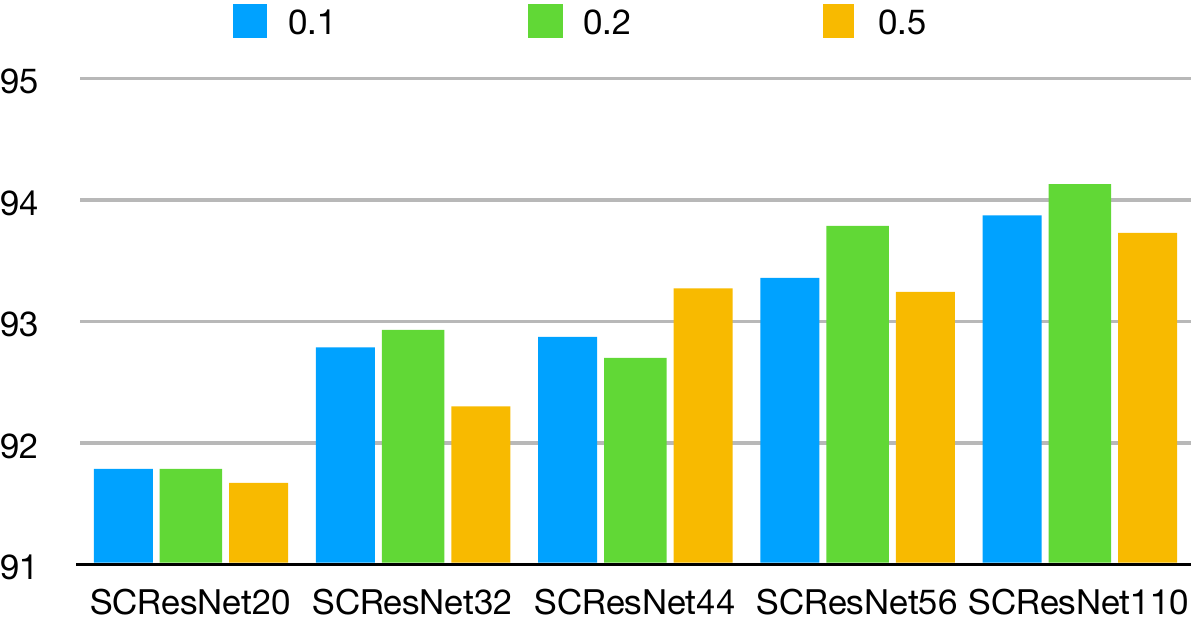} 
\caption{Different Weights of Short Circuit Connections.}
\label{FIG:comparison_param}
\end{figure}

\paragraph{Short Circuit Weights} It denotes the weight of SC connections, which is an adaptor of truncated sensitive from the rear layer to the front layers. The experiment results are summarized in Figure~\ref{FIG:comparison_param}. We observe that all the short circuit model performs well with SC weight $0.2$. If the weight increase to $0.5$, the performance decline more than the weight of $0.1$. This might cause by the gradient exist significant scale differences in two ends of short circuit connections. A higher weight ($0.5$) increases more fluctuation in gradient updates, leading to performance decline. So, the weight of short circuit connection default set to $0.2$ in our experiments. 

\begin{figure}[!htp]
\centering
\includegraphics[keepaspectratio,width=0.8\linewidth,height=6cm]{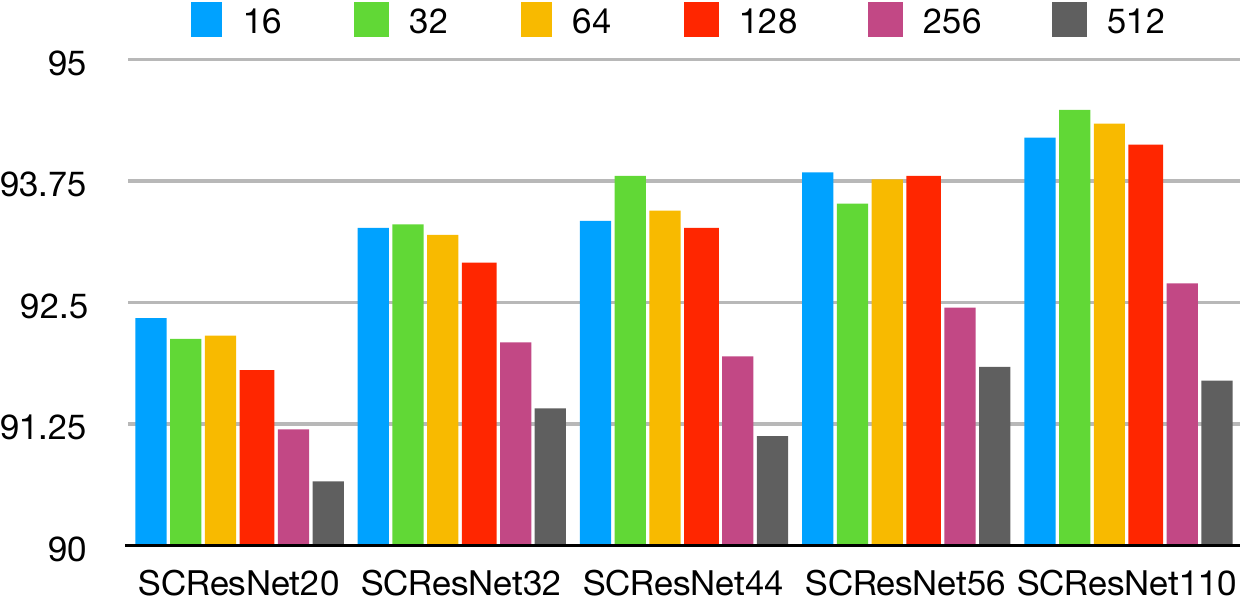} 
\caption{Different Batch Size of Short Circuit Connections.}
\label{FIG:comparison_batchsize}
\end{figure}

\paragraph{Batch Size} It is also a sensitive hyperparameter in gradient-based training tasks. Figure~\ref{FIG:comparison_batchsize} shows the batch size $32$ outperform others in most settings. With the batch size rising, the performance of short circuit models declines. The reason here we though is the update times also decrease in the same training epochs. Meanwhile, too small batch size also increases the gradient fluctuation training. So, the batch size sets $32$ in our SCResNet experiments. While, for the limitation of our GPU memory in BERT/Roberta experiments, we default set the batch size of $2$ in multiple choice question answering tasks.


\section{Conclusion}
In this work, we propose a novel method named short circuit to enhance the gradient learning in deep neural networks. Our short circuit introduced the gradient truncation of RNN into the shortcut of FNN which significant promote the gradient learning of the backbone feedforward neural networks. The experiments demonstrate the superiority of our method over the baselines on computer vision and natural language tasks. In the future, we plan to further apply our method to the larger datasets.

\section{References}
\bibliographystyle{elsarticle-num}
\bibliography{mybibfile}

\end{document}